\documentclass[letterpaper]{article} 
\usepackage{amsmath}
\usepackage{aaai25}  
\usepackage{times}  
\usepackage{helvet}  
\usepackage{courier}  
\usepackage[hyphens]{url}  
\usepackage{graphicx} 
\usepackage{natbib}
\usepackage{caption}
\usepackage{algorithm}
\usepackage{algorithmic}
\usepackage{indentfirst}
\usepackage{newfloat}
\usepackage{listings}
\DeclareCaptionStyle{ruled}{labelfont=normalfont,labelsep=colon,strut=off} 
\floatstyle{ruled}
\newfloat{listing}{tb}{lst}{}
\floatname{listing}{Listing}
\usepackage{subcaption}

\title{Active Learning and Explainable AI for \\Multi-Objective Optimization of Spin Coated Polymers}
\author{
    Brendan Young\textsuperscript{\rm 1}, Brendan Alvey\textsuperscript{\rm 1},
    Andreas Werbrouck\textsuperscript{\rm 2}, Willl Murphy\textsuperscript{\rm 2}, \\ James Keller\textsuperscript{\rm 1}, Matthias J. Young\textsuperscript{\rm 2,4}, Matthew Maschmann\textsuperscript{\rm 3,4}
}
\affiliations{
    \textsuperscript{\rm 1}Department of Electrical Engineering and Computer Science, University of Missouri, Columbia MO 65201 \\
    \textsuperscript{\rm 2}Department of Chemical and Biomedical Engineering, University of Missouri, Columbia MO 65201 \\
    \textsuperscript{\rm 3}Department of Mechanical and Aerospace Engineering, University of Missouri, Columbia MO 65201 \\
    \textsuperscript{\rm 4} MU Materials Science and Engineering Institute, University of Missouri, Columbia MO 65201 \\
    \{bmywzx, bma3md, awfk9, kellerj, mjym8d, maschmannm\}@umsystem.edu
}

\begin{document}

\maketitle

\begin{abstract}
Spin coating polymer thin films to achieve specific mechanical properties is inherently a multi-objective optimization problem. We present a framework that integrates an active Pareto front learning algorithm (PyePAL) with visualization and explainable AI techniques to optimize processing parameters. PyePAL uses Gaussian process models to predict objective values (hardness and elasticity) from the design variables (spin speed, dilution, and polymer mixture), guiding the adaptive selection of samples toward promising regions of the design space. To enable interpretable insights into the high-dimensional design space, we utilize UMAP (Uniform Manifold Approximation and Projection) for two-dimensional visualization of the Pareto front exploration. Additionally, we incorporate fuzzy linguistic summaries, which translate the learned relationships between process parameters and performance objectives into linguistic statements, thus enhancing the explainability and understanding of the optimization results. Experimental results demonstrate that our method efficiently identifies promising polymer designs, while the visual and linguistic explanations facilitate expert-driven analysis and knowledge discovery.

\end{abstract}

\section{Introduction}\label{sec:Intro}

Developing new materials is a challenging and time-consuming task. Traditionally, discovering new materials and optimizing synthesis procedures requires significant resources and long validation cycles \cite{MaterialScienceHard}. One application where these challenges are noticeable is spin coating.
Achieving an optimal spin-coated polymer film requires balancing multiple competing material properties. For example, both high hardness to resist deformation under load and low modulus of elasticity to remain flexible and to accommodate mechanical stress could be desired. Because materials design criteria are often contradictory, optimizing the resulting material becomes an extremely complex process, and global optimization is rarely possible. Instead, it is important to find design parameters that are non-dominated along particular criterion axes, the so-called Pareto front. 

To address these challenges, AI is a powerful tool that can be leveraged to help accelerate materials discovery and optimization. These AI-driven methods can facilitate the identification of promising material compositions and guide the experimental process through active learning techniques to find the Pareto front.  Bayesian optimization strategies \cite{3DPrinting}, such as $\epsilon$-Pareto Active Learning ($\epsilon$-PAL) \cite{ePAL, PyePAL}, have proven to be particularly effective in guiding experimental design by prioritizing the most informative combinations of parameters. However, when human researchers conduct these experiments, it is essential to provide transparent justifications for why certain parameter sets are selected and how it will benefit the materials science application.

An additional challenge in AI-driven materials research is the complexity of the design space, which often involves many interdependent parameters. This complexity can make it difficult for human scientists to interpret AI-generated recommendations and integrate them into the experimental workflow. Explainable Artificial Intelligence (XAI) techniques address this challenge by offering interpretable explanations that enhance trust and decision-making in AI-assisted research.

\begin{figure*}[!t]
\centering
\includegraphics[width=\textwidth]{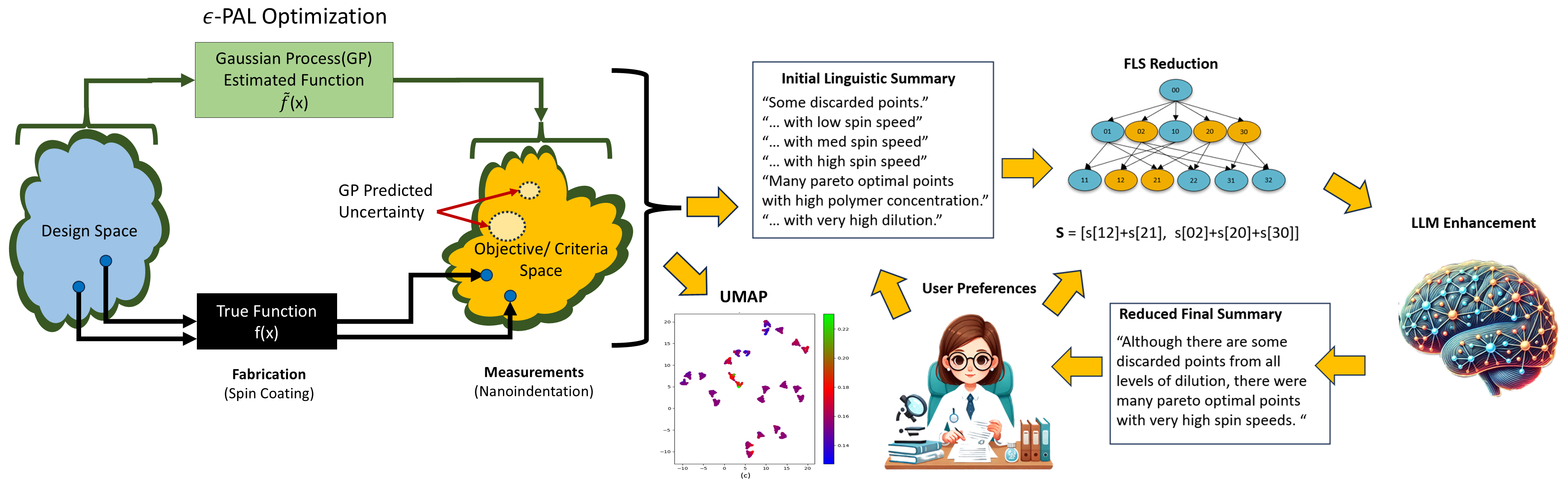}
\caption{On the left is the $\epsilon$-PAL workflow, where a Gaussian process model is iteratively updated to map the design space (e.g., spin coating parameters such as spin speed, dilution, and polymer concentration) to the objective space (e.g., nanoindentation measurements such as elasticity and hardness). Points are sampled to minimize the model's uncertainty. On the right, a FLS of the optimization data is constructed, evaluated, and simplified based on user preferences.}
\label{fig:overview}
\end{figure*}

In this paper, we explore two different XAI techniques, fuzzy linguistic summaries (FLSs) \cite{FuzzySummaries1} and Uniform Manifold Approximation and Projection (UMAP) \cite{UMAP} methods, to generate explanations that support experimental justification in the context of optimizing spin-coated polyvinylpyrrolidone (PVP) films. By providing insights into the relationships between spin-coating parameters and material properties, these methods help researchers understand and validate AI-driven choices. Furthermore, when explanations are insufficient or domain expertise suggests alternative strategies, scientists can use these insights to refine and redirect experiments toward more promising regions of the design space. Figure \ref{fig:overview} provides an overview of our process. 

The remainder of the paper is organized as follows: Section \ref{sec:Background} provides background information on $\epsilon$-PAL and its relevance to materials optimization. Section \ref{sec:XAI} describes the XAI techniques used to create explanations. Section \ref{sec:Experiment} presents the application of $\epsilon$-PAL and XAI techniques to the spin-coating experiment. Finally, Section \ref{sec:Conclusion} discusses the conclusions and future research directions.

\section{Background}\label{sec:Background}
\subsection{Related Work}
This work intersects multiple concepts such as multi-criteria decision making \cite{MCDM}, optimization under uncertainty \cite{OUC}, design of experiments \cite{DoE}, and integration of domain knowledge with machine learning models to improve decision making and efficiency in materials discovery. Active learning has been applied to other areas of materials discovery, such as the optimization of chemical reaction conditions \cite{ChemicalReactions}, the development of new battery materials \cite{Batteries}, and the design of alloys \cite{Alloys}. By combining these methodologies, AI can effectively navigate complex design spaces, identify optimal material compositions, and streamline the experimental process. Furthermore, these approaches enable the development of adaptive systems that learn from previous experiments and improve over time. It takes a combination of techniques such as these to overcome the numerous challenges in materials science, driving faster and more reliable advances in the creation of new materials with tailored properties.

\subsection{$\epsilon$-PAL}
An active learning technique designed for multi-objective optimization problems that we use is known as $\epsilon$-PAL \cite{ePAL}. More specifically we are using a python implementation PyePAL \cite{PyePAL}. Traditional exhaustive search methods to identify the Pareto-optimal set of design parameters are computationally expensive, making active learning approaches like $\epsilon$-PAL highly beneficial. The core idea of $\epsilon$-PAL is to efficiently sample the design space while ensuring that the selected solutions approximate the true Pareto front within a user-defined tolerance, denoted by $\epsilon$. This approach allows users to sacrifice a specified amount of accuracy in order to significantly reduce the number of required experiments compared to conventional methods.

At its core, $\epsilon$-PAL relies on Bayesian optimization modeling the mapping between design and criteria spaces as Gaussian processes (GPs). GPs are particularly useful in optimization settings where function evaluations are expensive or noisy, as they provide uncertainty estimates for unobserved points. The algorithm iteratively selects points where the model is most uncertain to evaluate next, focusing on those likely to be Pareto-optimal while discarding suboptimal candidates. The tolerance parameter $\epsilon$ defines how closely the predicted Pareto front must approximate the actual Pareto front; a larger $\epsilon$ allows for coarser approximations, reducing the number of evaluations needed, while a smaller $\epsilon$ leads to finer granularity at a higher experimental cost.

The execution of $\epsilon$-PAL involves several key stages. First, a Gaussian process regression model is trained on an initial set of function evaluations, providing estimates for the mean and variance of objective function values across the design space. The algorithm then proceeds with an iterative sampling process, identifying and evaluating the most uncertain or potentially optimal points. Points confidently deemed suboptimal are discarded, while others that may meet the $\epsilon$-Pareto criteria are retained. The process continues until the algorithm converges, meaning that no further points need to be evaluated to guarantee an $\epsilon$-accurate Pareto set with high probability.

One of the strengths of $\epsilon$-PAL is its theoretical guarantee on sample efficiency. The algorithm has a theorem that provides an upper bounds on the number of experiments required to achieve the desired level of accuracy. It does this by utilizing properties of Gaussian processes such as the kernel function's smoothness assumptions. These assumptions help to reduce the number of required samples, speeding up active learning exploration. Such aspects make it particularly suitable for applications where function evaluations are costly, such as hardware design, software optimization, and engineering design spaces.

\subsubsection{Example}

\begin{figure}
    \centering
    \subfloat[\centering]{
        \includegraphics[width=0.5\linewidth]{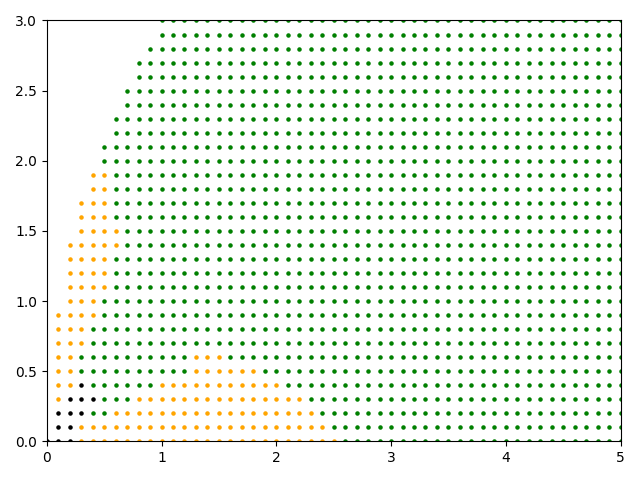}
    }
    \subfloat[\centering]{
        \includegraphics[width=0.5\linewidth]{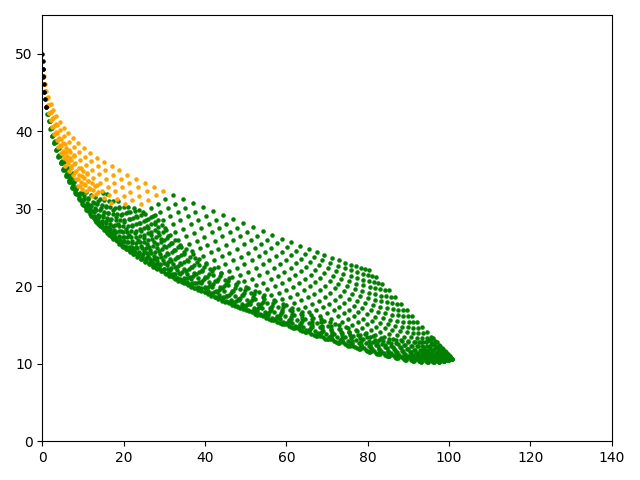}
    }
    \caption{Iteration 7 of PyePAL on the Binh-Korn benchmark. (a) Design Space; (b) Criteria Space.  The few black points are those identified as Pareto optimal, orange points are those currently discarded, and green points are unknown.}
    \label{fig:binh_korn_iteration_6}
\end{figure}
\begin{figure}
    \centering
    \subfloat[\centering]{
        \includegraphics[width=0.5\linewidth]{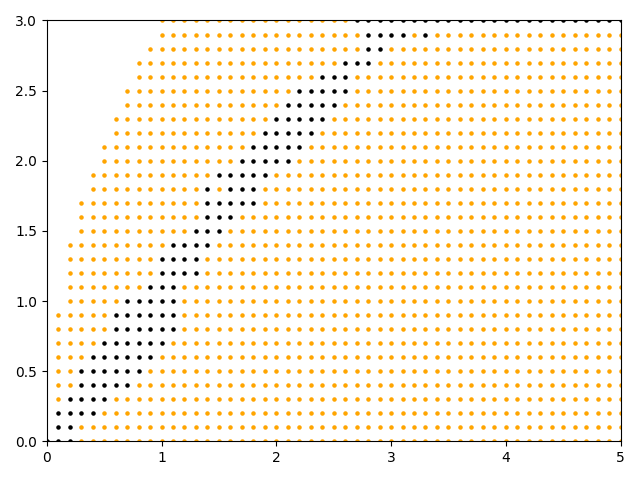}
    }
    \subfloat[\centering]{
        \includegraphics[width=0.5\linewidth]{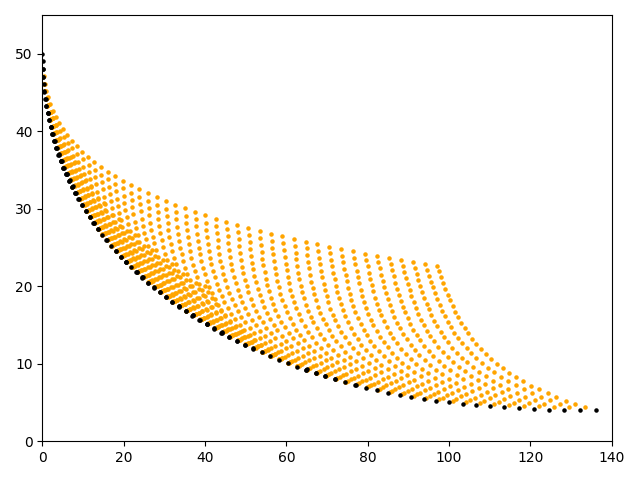}
    }
    \caption{Iteration 14 of PyePAL on the Binh-Korn benchmark. (a) Design Space; (b) Criteria Space.  The complete Pareto front is identified in black, orange points are those currently discarded, and there are no undecided points.}
    \label{fig:binh_korn_iteration_14}
\end{figure}

To illustrate the effectiveness of $\epsilon$-PAL, we apply it to the Binh-Korn multi-objective optimization problem, a well-known benchmark in multi-objective optimization. The Binh-Korn problem involves two conflicting objective functions constrained within a defined input space, making it an ideal test case for evaluating optimization algorithms. Its Pareto front is well-documented, clearly comparing the algorithm's predictions and the optimal solutions. For this example, we are searching for a $\epsilon$-Pareto Front where $\epsilon = 0.01$

Our demonstration consists of two sets of figures, capturing different stages of the optimization process.

At Iteration 7, the algorithm has sampled eight points, some of which have been classified as Pareto-optimal, while others have been identified as suboptimal and discarded. In Figure \ref{fig:binh_korn_iteration_6}a, the Pareto-optimal points are clustered in the bottom-left region, while in Figure \ref{fig:binh_korn_iteration_6}b, they appear in the top-left area of the criterion space. The predicted values at this stage are beginning to approximate the true criterion space, but significant uncertainty remains. Notably, in Figure \ref{fig:binh_korn_iteration_6}b, the regions that exhibit the most significant inaccuracies also correspond to the areas of highest uncertainty, reflecting the algorithm's exploration focus. 

By the final iteration (Iteration 14, with 15 sampled points), the algorithm has fully classified all points as either Pareto-optimal or suboptimal. In Figure \ref{fig:binh_korn_iteration_14}b, the predicted criterion space closely matches the underlying criterion space, and the determined Pareto front accurately reflects the Pareto front. These results demonstrate the efficiency of $\epsilon$-PAL in refining the search space while minimizing function evaluations.

If this were an actual materials discovery problem, the next steps could involve generating explanations for the selected Pareto-optimal values or further approximating the surface estimated by the Gaussian Process (GP) model. These insights can help guide future experimental efforts, improving efficiency and interpretability in the materials optimization process.

\section{Explainable AI Techniques}\label{sec:XAI}

\subsection{Fuzzy Linguistic Summaries}

Although large language models (LLMs) have become incredible popular over the past few years with impressive products such as Open AI's ChatGPT \cite{openai2023gpt4}, they cannot reliably analyze large datasets and often hallucinate answers \cite{nelson2024needlehaystackmemorybased, sriramanan2024llmcheck}. A more reliable family of techniques for producing accurate linguistic descriptions of data is known as fuzzy linguistic summaries (FLS). They can be useful for many things, including analyzing complex models that operate essentially as black boxes \cite{10195035} (such as neural networks), for summarizing databases, \cite{4358814}, explanations of time-series \cite{5364881}, and anomaly detection \cite{6622481}. Especially when considering several different variables at once, it can become nearly impossible to easily visualize the relationships between different combinations of conditions. This is where FLS excels, allowing for systematic evaluation of these combinations, determining which relationships hold true, and presenting them to the end-user. 

FLS are constructed from a number of linguistic statements that describe data in a human-comprehendible, structured manner. The set of these statements, also known as linguistic protoforms, can be generated systematically based on all of the combinations of attributes and fuzzy predicates that describe those attributes. One form these statements may take is, \textit{``Of the Ys that are P, Q are R"}, where:

\begin{itemize}
    \item $Y$ represents objects in the dataset, such as points on the Pareto front.
    \item $P$ is the summarizer, consisting of attribute-predicate pairs, in our case describing the design space (e.g. \textit{``low spin speed and high dilution"})
    \item $Q$ is a quantifier indicating the prevalence of data, such as ``Few'' or ``Many.''
    \item $R$ is a qualifier describing a quality metric, such as a measure of uncertainty, with predicates like ``Low'' or ``High.''
\end{itemize}

To compute the membership for a given data sample $n$ to a particular summarizer:
\begin{align}
    \mu_{n}^P &= \bigwedge_{k,j \in S} \mu_{k,j}(x_n), \label{equa:summarizer}\\
    \mu_{n}^R &= \mu^{R}(r_{n}), \label{equa:qualifier}
\end{align}
where $k,j \in s$ refers to the fuzzy predicates in the summarizer of the statement, $S$, and $r_n$ is the attribute used by the qualifier (e.g. predictive uncertainty). In all equations, $\wedge$ is the chosen t-norm (we used the min operator). 

For each linguistic statement, various metrics can be computed. For example, truth (also known as validity) is a common metric which attempts to calculate how accurate a statement is with respect to a particular dataset. As noted in \cite{Alvey2025_FUZZ_IEEE} and \cite{Alvey2025_CAI}, one must be careful to choose protoform structures which precisely corresponds to their respective metrics so that the values computed are meaningful. As such, we use the following equations for calculating truth values: 
\begin{equation}
  T(s) = \mu^{Q}\!\Biggl(\frac{\sum_{n=1}^{N} \bigl(\mu_{n}^R \,\wedge\, \mu_{n}^P\bigr)}{D}\Biggr),
  \label{equa:truth_1}
\end{equation}
where $D$ is commonly $\sum_{n=1}^{N} \mu_{n}^P$, so the qualifier is membership function, $\mu^Q$ is evaluated on the ratio of samples which match the qualifier \textit{and} summarizer to those that match the summarizer. If the statement does not have a qualifier then $D$ is set to $N$ which could be the total number of samples in the dataset, or the number of samples of a specific category if we want our qualifiers and prototypes to reflect quantities relative to only a specific category (e.g. Pareto optimal points). We can assume $\mu^R=1$ in this case. Likewise, in the absence of a summarizer, we can assume $\mu^P=1$ and let $D$ be set to $N$. 

These summaries can often result in many statements with high truth values that contain highly redundant information. For example, we might find that a polymer has near optimal material properties under a large number of different conditions. In this case, each condition, and combination of conditions would produce linguistic statements that evaluate to having high truth values. To address this redundancy we use a FLS simplification technique first presented in \cite{Alvey2025_FUZZ_IEEE}. This method involves representing the set of possible linguistic statements as a hierarchical directed acyclic graph which can be processed to determine which statements actually need to be reported. A user-defined threshold is employed to initially remove statements which do not have a high truth value. After the graph processing, the resulting intermediate FLS is passed to an LLM for further enhancement. This way, we take advantage of LLMs in a task they have been proven to be reliable in (summarizing and formatting), while leaving the bulk of information processing to a principled heuristic. This combination of modern AI and reliable pruning of the FLS results in preservation of important information presented in a much more user-friendly form, improving end-user insights and knowledge discovery. 

\subsection{UMAP}
\begin{figure}
    \centering
    \includegraphics[width=1.0\linewidth]{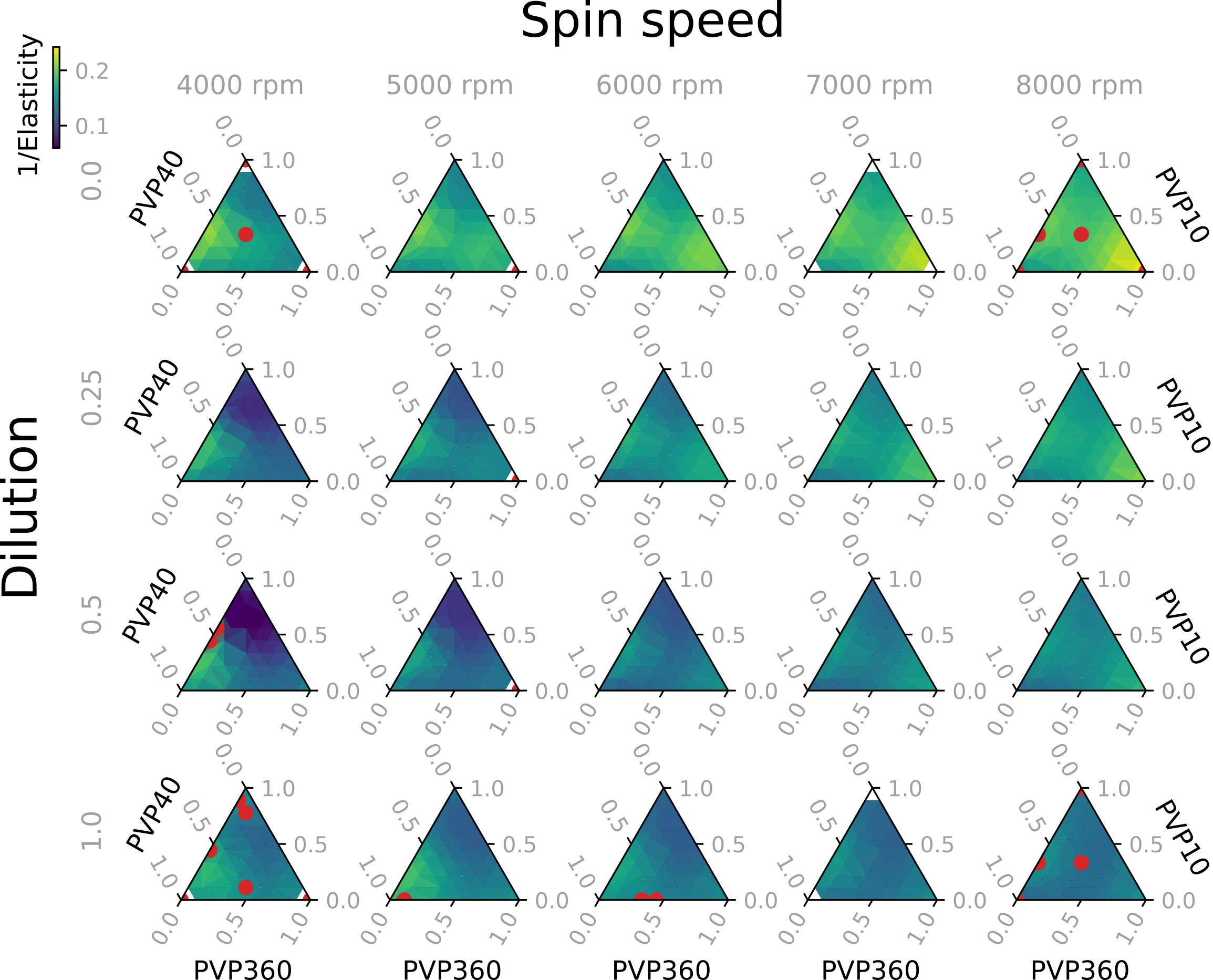}
    \caption{Human expert-developed projection of the spin coating output space of (inverse) elasticity. 5 input parameters are compressed in a 2D figure. Making such visualizations costs a lot of time and will ultimately fail when more experimental parameters are added. A more flexible approach of projecting the space is necessary.}
    \label{fig:InputSpace}
\end{figure}

In materials science, the design space often extends beyond three dimensions, making it difficult to visualize. High-dimensional datasets pose challenges in visualization, interpretation and analysis (Figure \ref{fig:InputSpace}), necessitating the use of dimensionality reduction techniques to represent the data in a more comprehensible form. One effective method for visualizing high-dimensional data is to employ dimensionality reduction techniques, which project the data into lower-dimensional spaces while preserving meaningful geometric relationships.

Uniform Manifold Approximation and Projection (UMAP) is a non-linear dimensionality reduction technique designed to preserve both global and local structures in high-dimensional data \cite{UMAP}.  UMAP is grounded in manifold learning and graph theory, making it well-suited for visualizing complex datasets and uncovering underlying structures.

UMAP constructs a weighted k-nearest neighbor (k-NN) graph from the high-dimensional input data. It uses this to model the local relationships between data points. The algorithm then uses the k-NN graph to create a low-dimensional embedding by minimizing a cost function that prioritizes local neighborhoods and global structure. This process is computationally efficient and balances local density preservation with global topology retention. UMAP offers a good balance between preserving local and global structures, computational efficiency, and scalability to large datasets.

A key advantage of UMAP is its ability to retain global relationships effectively, making it particularly useful for clustering and exploratory data analysis. UMAP is also inherently non-linear, enabling it to capture complex, non-Euclidean structures. Additionally, UMAP can utilize custom distance metrics and has tunable parameters  such as the number of neighbors and the minimum spanning tree connectivity. This makes UMAP highly adaptable to a wide range of applications, including bioinformatics, image analysis, and natural language processing.

Due to its efficiency and versatility, UMAP has gained widespread adoption in various scientific domains, particularly in the analysis of large-scale, high-dimensional datasets where structure-preserving dimensionality reduction is critical.

\section{Experiment}\label{sec:Experiment}
\begin{figure}
    \centering
    \includegraphics[width=1.0\linewidth]{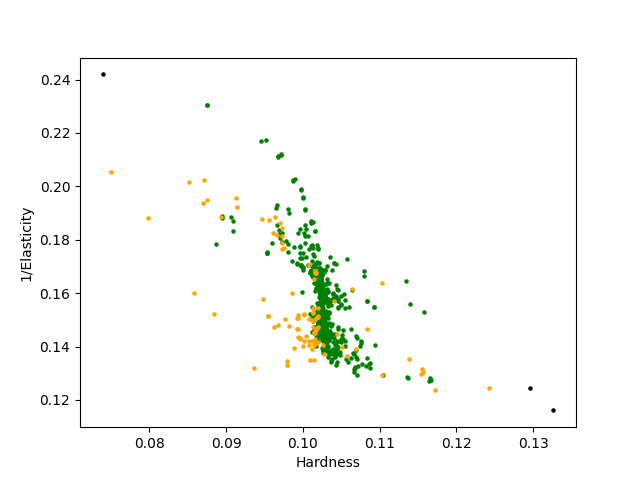}
    \caption{Optimization data from PyePAL after the 5th iteration. The pareto front has not fully been identified, but FLS statements can already be made about the model that guides the algorithm.}
    \label{fig:OutputSpace}
\end{figure}

To illustrate the application of XAI in materials science, we consider the process of spin-coating a polymer onto a substrate as described in Section \ref{sec:Intro}. The spin coating design space has many variables: the chemical structure of the monomer/polymer, molecular weight of the polymer (how long the chain of monomers is), polymer dilution, choice of solvent, spin acceleration, spin speed and spinning time. These all will impact the thickness, mechanical and optical properties of the film \cite{Lawrence1988}.

For the purpose of this research, the design space was limited to five parameters. Polyvinylpyrrolidone (PVP) was selected as a polymer. Three molecular weights (chain lengths) were picked: 10.000 (PVP10), 40.000 (PVP40) and 360.000 (PVP360). Typically shorter chain lengths lead to greater plasticity, but mixing these could lead to the kind of non-linear effects that could be interesting from a multiobjective optimization perspective. The polymers come in powder form, so stock solutions were prepared with ethanol. These stock solutions could be mixed in various concentrations $c$, adding a constraint $c_{\text{PVP10}} + c_{\text{PVP40}} + c_{\text{PVP360}} = 1$ (3 parameters). Such mixture can then further be diluted with more ethanol, from not diluted ($d=0$ to adding the same amount of ethanol ($d=1$)) (1 parameter). Finally, the spin speed was also considered (1 parameter, $S$). This leads to the coordinates used below ($c_{\text{PVP10}}$, $c_{\text{PVP40}}$, $c_{\text{PVP360}}$, $S$, $d$). All other parameters were kept the same.

The output space was limited to the mechanical properties (hardness and elasticity) of the spin-coated PVP. To characterize these, a small, hard tip is repeatedly pressed into and retracted from a substrate with increasing load (nanoindentation). From the load/displacement curve of this compound measurement, (nano)mechanical properties of the material of interest can be derived. However, hardness —resisting deformation under applied pressure— and elasticity -allowing the material to return to its original state after deformation- are inherently correlated in first order. By inverting the elasticity, effectively anti-correlate materials properties, and the second-order effects allow us to define two Pareto fronts: points that, for the same hardness, have higher-than-than-expected elasticity, and points that, for the same hardness, have lower-than-expected elasticity. Here, we chose to optimize a coating with both high hardness and low modulus of elasticity (high inverse elasticity). 

This presents a complex, multi-objective optimization problem well-suited for XAI techniques, which can provide insights into the relationships between processing parameters and material properties, ultimately guiding the development of optimized coatings. Every iteration, Pyepal was used to generate the next 3 points of interest, which were subsequently fabricated (the desired formulation and dilution was prepared and deposited at the requested revolutions per minute) and tested with the nanoindenter by a human researcher.

Figure \ref{fig:InputSpace} illustrates a human-designed visualization of the input domain consisting of a grid of ternary composition diagrams, where each row represents a specific dilution, each column represents a specific spin speed, and each triangle is a ternary composition diagram of the three molecular weights: PVP10, PVP40, and PVP360. In this figure, the triangles are colored based on the estimated elasticity of a given combination of the five parameters. The red points in the figure indicate the specific parameter sets that have been experimentally sampled.

Figure \ref{fig:OutputSpace} presents the estimated values of both the hardness and the inverse elastic modulus for the sampled points. The black points represent the parameter sets that have been determined to be part of the $\epsilon$-Pareto Front, indicating optimal trade-offs between hardness and elasticity. The orange points correspond to parameter sets that have been determined not to be on the Pareto front, meaning they are suboptimal compared to other configurations. The green points indicate cases where uncertainty in the measurements is large enough that they cannot yet be definitively classified as either Pareto-optimal or suboptimal. Although the quantity of experimental data is insufficient to establish a reliable model, employing linguistic summaries using FLS to describe the current state of the experimental data yields interesting and valuable insights into process-property relationships.

FLS were produced by analyzing the optimization data and evaluating the truth value for each possible statement. An example intermmediate statement that the FLS produced before it was passed to the LLM for futher simplification was, \textit{``Of the design points from very large pvp360 concentration, very large spin speed, medium dilution, some are pareto optimal points."} We used a threshold of $0.95$ to remove statements that did not have a high truth value. Our membership functions were chosen to evenly divide our design space and sum to $1$. A visualization of each of our membership functions is found in Figure \ref{fig:fuzzy_membership_functions}. Below are the results of the final FLS for this iteration:

\begin{figure}
    \centering
    \includegraphics[width=1.0\linewidth]{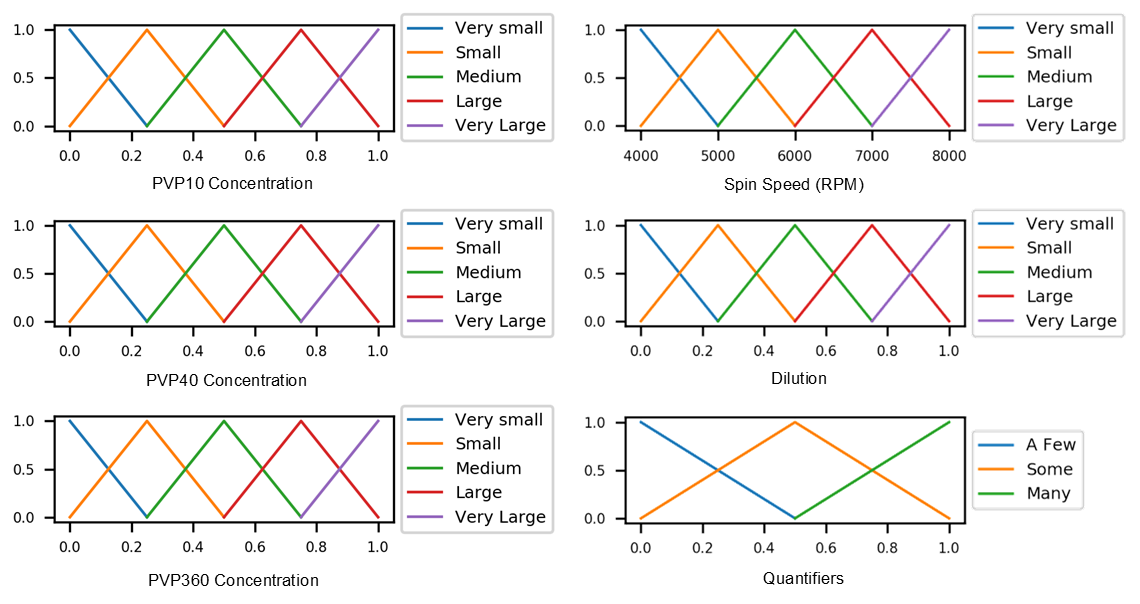}
    \caption{Fuzzy membership functions. Each of the input variables (polymer concentrations, spin speed, and dilution) form the summarizer. The quantifier is evaluated on ratios of memberships according to equation \ref{equa:truth_1}}
    \label{fig:fuzzy_membership_functions}
\end{figure}

\begin{itemize}
    \item \textbf{Few Pareto Optimal Points:}
    \begin{itemize}
        \item Medium concentrations of pvp10, pvp40, and pvp360 combined with very large spin speeds and dilutions.
    \end{itemize}
    \item \textbf{Some Optimal Points:}
    \begin{itemize}
        \item Small pvp10 and pvp40 with very large pvp360, very large spin speeds, and small dilution.
        \item Very large pvp10 with very large spin speeds and very small dilution.
        \item Very large pvp360 with very large spin speeds and medium dilution.
    \end{itemize}
    \item \textbf{Many Discarded Points:}
    \begin{itemize}
        \item Large pvp10 with medium pvp40 and large or very large dilutions.
        \item Medium pvp10 with large pvp40 and very large spin speeds across various dilutions.
    \end{itemize}
    \item \textbf{Some and Few Discarded Points:}
    \begin{itemize}
        \item Large pvp10 with very large spin speeds and dilutions.
        \item Medium pvp40 with very large spin speeds and dilutions.
        \item Specific combinations involving large pvp360 and small pvp40 concentrations.
    \end{itemize}
    \item \textbf{Many Undecided Points:}
    \begin{itemize}
        \item Diverse combinations of pvp10, pvp40, and pvp360 with varying spin speeds and dilution levels.
        \item Particularly prevalent with medium pvp360 concentrations and differing spin speeds and dilutions.
    \end{itemize}
    \item \textbf{Some and Few Undecided Points:}
    \begin{itemize}
        \item Large pvp40 with very small spin speeds and dilutions.
        \item Small pvp10 with various dilution and spin speed combinations.
        \item Specific high-concentration scenarios with varying spin speeds.
    \end{itemize}
\end{itemize}

\vspace{1em}

\textbf{\large Overall Insights}
\begin{itemize}
    \item \textbf{Pareto Optimality} is rare, achievable only under specific medium to high concentration and dilution settings combined with very large spin speeds.
    \item \textbf{Discarded Points} are frequent in high concentration and dilution environments, particularly with large pvp10 and pvp40 levels.
    \item The majority of design points fall into the \textbf{Undecided} category, indicating a need for further analysis or optimization under varied parameter settings.
\end{itemize}

\begin{figure*}
    \centering
    \includegraphics[width=1.0\linewidth]{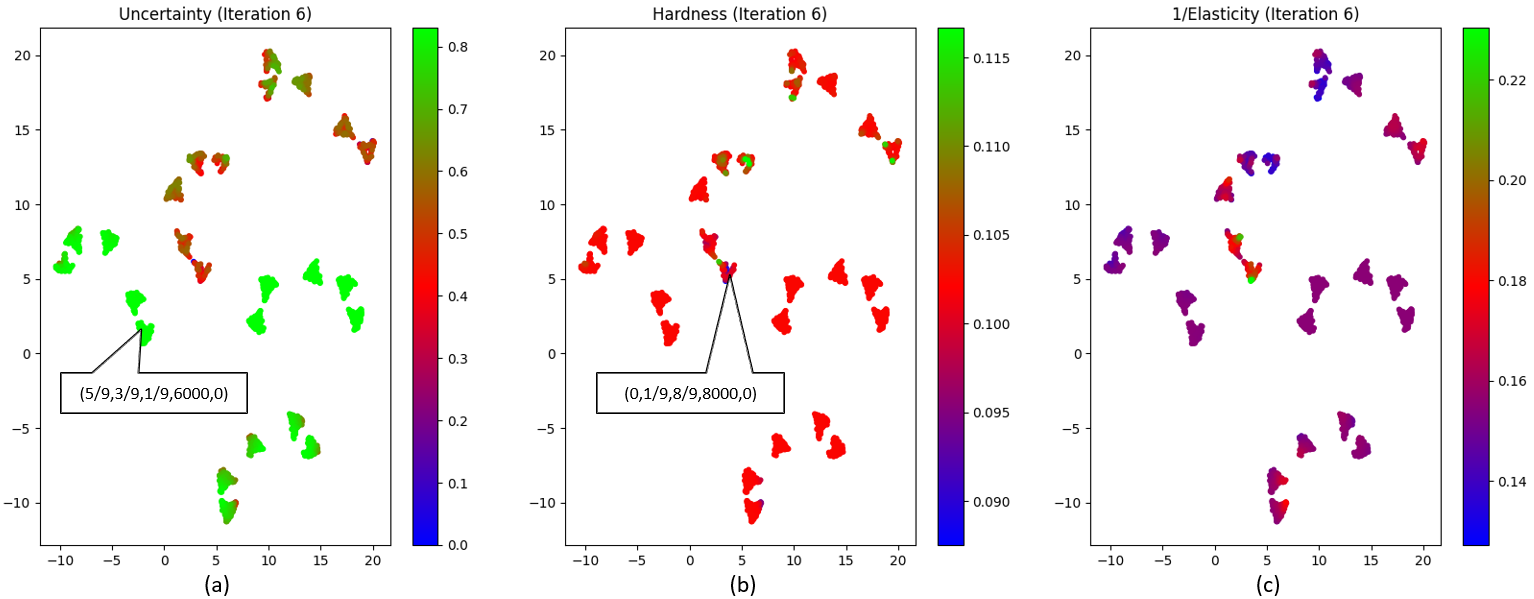}
    \caption{UMAP projection of the design space. (a) Points colored by uncertainty, highlighting unexplored regions. (b) Points colored by hardness, showing the distribution of material properties. (c) Points colored by inverse modulus of elasticity, illustrating the correlation between design parameters and mechanical performance.} 
    \label{fig:UMAP}
\end{figure*}

From an expert point of view, these FLS make sense. High spin speeds (and low dilutions) will produce thinner films than other conditions. These films will be too thin to measure reliably with the nanoindenter, which will start to pick up signal from the substrates underneath. While the model will converge towards a good prediction of the nanoindentation values regardless of the film stack, care should be taken to avoid systematic errors like this early in the process.

Following the FLSs, we next study a UMAP projection of the input space to generate a visual explanation of what has been explored, as shown in Figure \ref{fig:UMAP}. This figure consists of three charts. Figure \ref{fig:UMAP}a displays the UMAP projection colored by uncertainty. Figure \ref{fig:UMAP}b presents the UMAP projection colored by hardness. Figure \ref{fig:UMAP}c shows the UMAP projection colored by the inverse of the modulus of elasticity.

More generally, UMAP preserves interesting structural patterns within the input space. Notably, the projection forms 25 distinct triangular shapes, which directly correspond to the number of parameter combinations, as the input space consists of five dilution levels and five spin speeds independent of the PVP mixture composition. This suggests that UMAP successfully captures the interdependency between mixture variables, preserving meaningful structure in the reduced-dimensional space.

Additionally, UMAP groups the triangles into clusters of five, this corresponds to distinct spin speed values. This structured organization shows how UMAP naturally identifies and preserves key relationships within the dataset, making its projection remarkably similar to the human-designed visualization. This suggests that even in the absence of explicit human design, UMAP can generate interpretable and coherent visual representations of high-dimensional materials science design spaces.

Further insights can be drawn from Figure \ref{fig:UMAP}. For instance, the $\epsilon$-PAL algorithm suggests an optimal point for the next experiment, corresponding to the parameter set (5/9, 3/9, 1/9, 6000, 0), where the format represents (PVP10, PVP40, PVP360, spin speed, dilution). The UMAP projection of this point is highlighted in Figure \ref{fig:UMAP}a. At this stage, a human scientist, relying on their experience with the domain, can inspect the UMAP to evaluate which parts of the input space are likely to be affected by completing an experiment at these parameters.

In this case, Figure \ref{fig:UMAP}a shows that the suggested point is surrounded by points with high uncertainty. This suggests that performing an experiment at this point could be valuable for reducing uncertainty and refining the model's understanding of the input-output relationships.

Alternatively, suppose a scientist believes that the suggested experiment will not provide a beneficial result or is experimentally unattainable. For example, if low dilutions have been observed to produce poor film uniformity. In this case, the scientist can use the UMAP projection along with their domain expertise to identify a potentially more promising experimental point. For instance, (0, 1/9, 8/9, 8000, 0)—visualized in Figure \ref{fig:UMAP}b—represents another candidate. This point exhibits mid-range hardness but high elasticity, making it an interesting alternative depending on the desired material properties. We note that PyePAL provides the functionality to suggest multiple experiments, which can be plotted in a UMAP projection, prompting a user to select their next experiment from among these options using both the model predictions and their domain knowledge

\section{Conclusion}\label{sec:Conclusion}

In this work, we explored the integration of XAI techniques into the multi-objective optimization of spin-coated polymer films, demonstrating how AI-driven methodologies can enhance materials discovery while maintaining interpretability. We employed $\epsilon$-PAL to efficiently navigate the high-dimensional design space and identify optimal trade-offs between hardness and elasticity. To improve interpretability, we utilized FLS and UMAP visualization to provide justifications for experimental selections. Our results show that $\epsilon$-PAL reduces experimental evaluations while maintaining accuracy, FLSs help researchers identify key trends, and UMAP preserves structure in high-dimensional spaces, enabling intuitive visualizations.

Future work could focus on applying these XAI techniques to other materials science problems, enhancing the human interpretability of the XAI outputs, and exploring the potential of replacing a human user with a LLM agent trained on subject-matter literature. While our FLSs help answer \textit{what} is occurring during optimization, we aim to investigate \textit{why} certain trends emerge. LLMs could be used to suggest possible explanations for the relationships identified by the FLSs. Additionally, we plan to advance our techniques by adding more expressiveness to the FLSs, enabling them to describe how optimization data evolve with each iteration and identify which design parameters result in specific output criteria. Another promising direction for future research is the possibility of allowing user intervention during the optimization process. If an expert notices a trend or suspects a beneficial design point during an iteration, they could direct the experiment to explore alternatives to the recommendations provided by PyePAL. This could be facilitated through an interactive tool, creating a unique collaboration between human expertise and AI to further accelerate the pace of knowledge discovery. 

\section*{Acknowledgments}
This material is based upon work supported by the Engineering Research and Development Center - Information Technology Laboratory (ERDC-ITL) under Contract No. W912HZ24C0022

\bibliography{aaai25}
\end{document}